\documentclass[10pt,twocolumn,letterpaper]{article}

\PassOptionsToPackage{switch}{lineno}
\usepackage{cvpr}      %

\newcommand{\todo}[1]{{#1}}
\newcommand{\TODO}[1]{\textbf{\color{red}[TODO: #1]}}

\renewcommand{\TODO}[1]{}
\renewcommand{\todo}[1]{}

\newcommand{\canskip}[1]{#1}

\definecolor{darkgreen}{rgb}{0,0.5,0}

\newcommand{\myparagraph}[1]{\textbf{#1}~} %

\usepackage[accsupp]{axessibility}  %

\usepackage{multirow}
\usepackage[export]{adjustbox}

\usepackage{algpseudocode}
\usepackage{algorithm}
\usepackage{tabularx}
\usepackage{amsmath}
\usepackage{tabularray}
\makeatletter
\newcommand{\multiline}[1]{%
  \begin{tabularx}{\dimexpr\linewidth-\ALG@thistlm}[t]{@{}X@{}}
    #1
  \end{tabularx}
}
\makeatother

\newcommand{\topk}[1]{\mathop{\mathrm{top}^{#1}}}

\newcommand{\std}[1]{ \ensuremath{\pm{}}#1}

\def\tablescaler{0.8}

\definecolor{cvprblue}{rgb}{0.21,0.49,0.74}
\usepackage[pagebackref,breaklinks,colorlinks,allcolors=cvprblue]{hyperref}

\def\paperID{20939} %
\def\confName{CVPR}
\def\confYear{2026}

\title{RoaD: Rollouts as Demonstrations for Closed-Loop Supervised Fine-Tuning of Autonomous Driving Policies}

\author{
Guillermo Garcia-Cobo$^{*1}$ \quad
Maximilian Igl$^{*1}$ \quad
Peter Karkus$^{*1}$ \quad
Zhejun Zhang$^{*2\dagger}$ \\[2pt]
Michael Watson$^{1}$ \quad
Yuxiao Chen$^{1}$ \quad
Boris Ivanovic$^{1}$ \quad
Marco Pavone$^{1,3}$ \\[3pt]
$^{1}$NVIDIA Research \quad 
$^{2}$Huawei VN Research Center \quad 
$^{3}$Stanford University\\[2pt]
{\tt\small \{guillermog, migl, pkarkus\}@nvidia.com}
}

\begin{document}
\maketitle

\begin{abstract}

Autonomous driving policies are typically trained via open-loop behavior cloning of human demonstrations. However, such policies suffer from covariate shift when deployed in closed loop, leading to compounding errors.
We introduce Rollouts as Demonstrations (RoaD), a simple and efficient method to mitigate covariate shift by leveraging the policy’s own closed-loop rollouts as additional training data.
During rollout generation, RoaD incorporates expert guidance to bias trajectories toward high-quality behavior, producing informative yet realistic demonstrations for fine-tuning.
This approach enables robust closed-loop adaptation with orders of magnitude less data than reinforcement learning, and avoids restrictive assumptions of prior closed-loop supervised fine-tuning (CL-SFT) methods, allowing broader applications domains including end-to-end driving.
We demonstrate the effectiveness of RoaD on WOSAC, a large-scale traffic simulation benchmark, where it performs similar or better than the prior CL-SFT method; and in AlpaSim, a high-fidelity neural reconstruction–based simulator for end-to-end driving, where it improves driving score by 41\% and reduces collisions by 54\%.
\end{abstract}

\section{Introduction}
\label{sec:intro}

\begingroup
\renewcommand\thefootnote{*}
\footnotetext{Equal contribution, alphabetically sorted.}
\endgroup

\begingroup
\renewcommand\thefootnote{$\dagger$} %
\footnotetext{Work performed during an internship at NVIDIA Research.}
\endgroup

Autonomous vehicle (AV) policies are typically trained with behavior cloning
(BC) of human demonstrations, which is scalable but inherently open loop: it
assumes i.i.d. inputs and optimizes one-step accuracy under the dataset
distribution. Deployed in closed loop, policies influence their own
observations, creating a train-test mismatch that induces covariate shift,
compounds errors, and reduces robustness to long-tail and interactive scenarios.

On the other hand, End-to-end (E2E) policies are becoming the new norm of AV policy learning, 
which map sensor inputs directly to trajectories or
controls. By coupling perception,
prediction, and planning, they offer data efficiency, simpler deployment, and
better long-horizon coordination than hand-engineered
stacks~\citep{chen2024end,hwang2024emma,tian2024drivevlm,wayve2025,NVIDIASafety2025,wang2025alpamayo}.

While RL directly optimizes closed-loop behavior, it remains impractical for
end-to-end driving~\cite{kiran2021deep} due to brittle reward design and
the cost of safe exploration and high-fidelity simulation. This leaves a gap for
a scalable closed-loop training recipe for E2E driving that retains supervised simplicity and
data efficiency.

\begin{figure}[t]
  \centering
  \begin{subfigure}[t]{0.47\linewidth}
    \centering
    \includegraphics[width=\linewidth]{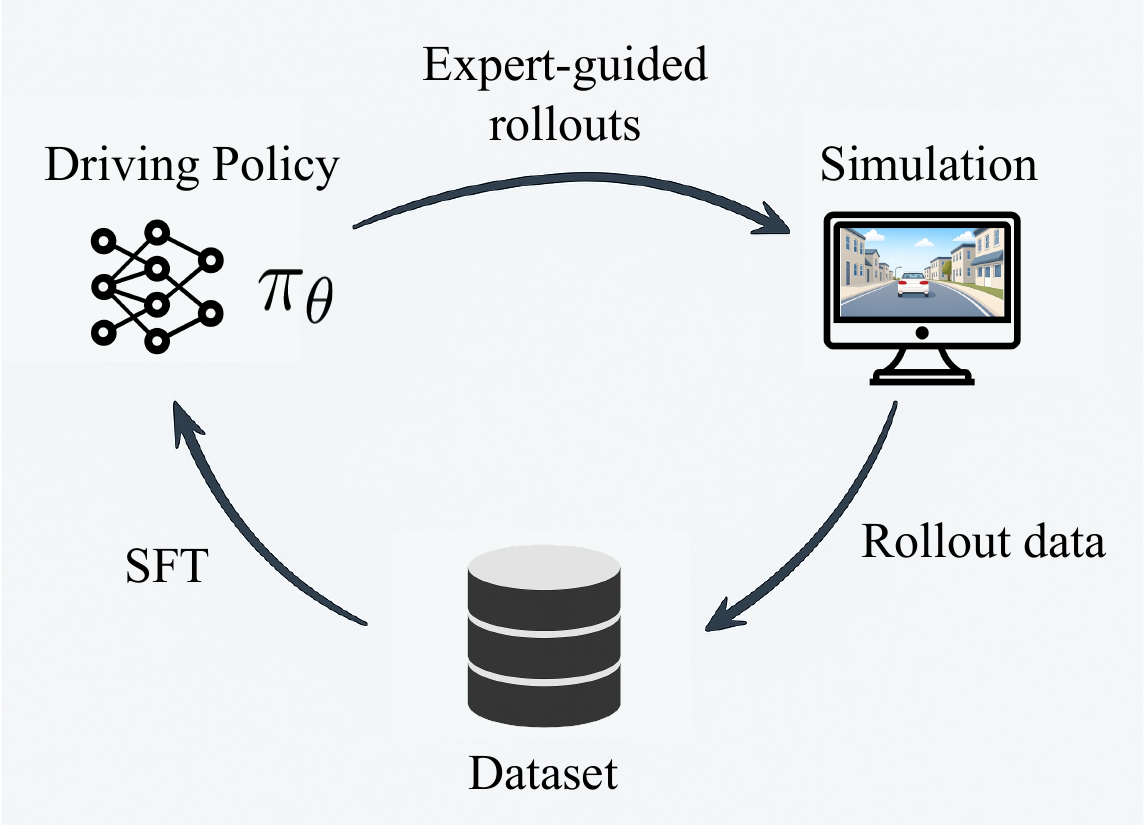}
  \end{subfigure}\hfill
  \begin{subfigure}[t]{0.5\linewidth}
    \centering
    \includegraphics[width=\linewidth]{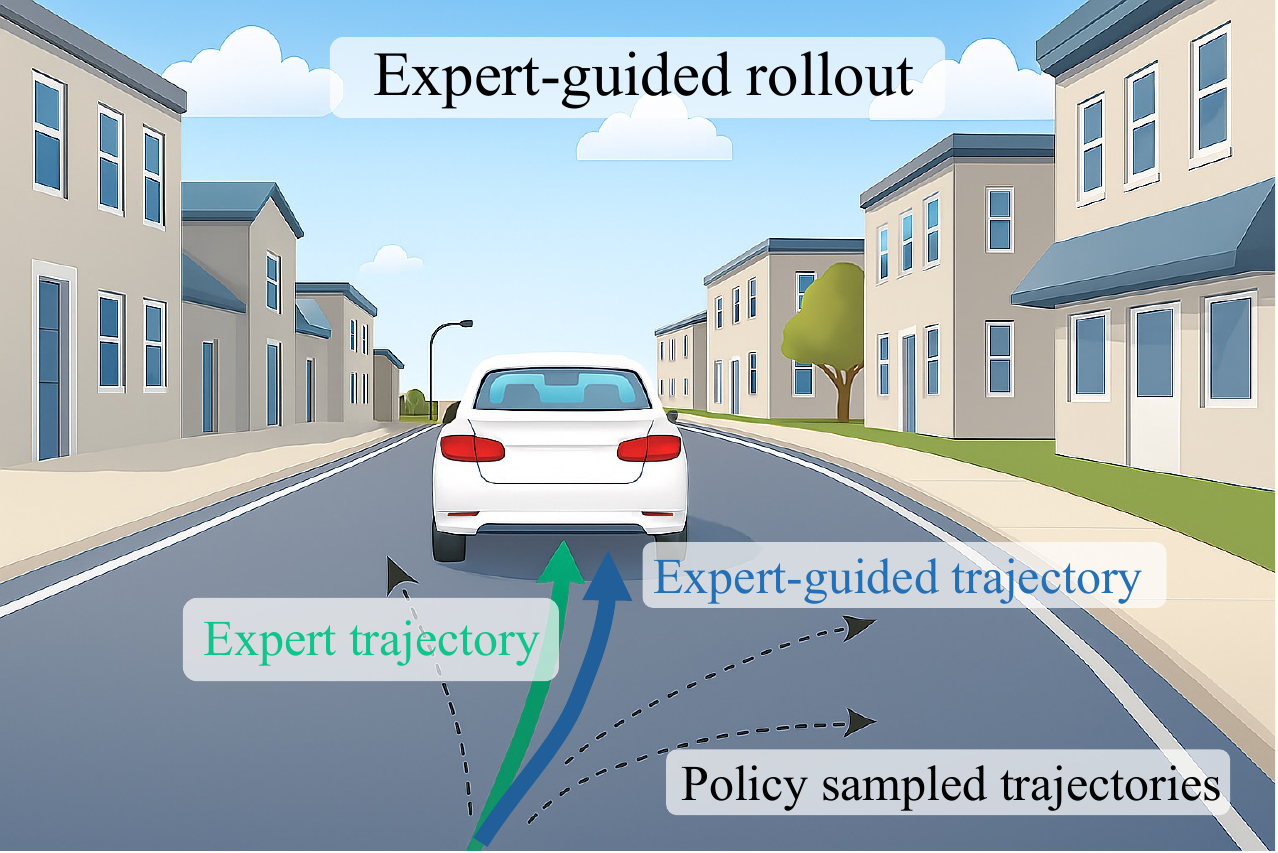} %
  \end{subfigure}
  \caption{\textbf{RoaD closed-loop SFT with expert-guided rollout}. \textbf{Left:} Expert-guided rollouts in simulation yield additional training data that is incorporated into the dataset and used to fine-tune the policy, improving subsequent rollouts.
  \textbf{Right:} Expert-guided trajectories are sampled from the policy and biased toward the expert to produce higher-quality demonstrations.}
  \label{fig:intro_road_cycle}
  \vspace{-1ex}
\end{figure}

Closed-loop supervised fine-tuning (CL-SFT) has recently emerged as a promising
alternative to RL (see \cref{fig:intro_road_cycle}). The core idea of CL-SFT is to generate expert-biased on-policy
rollouts in simulation and use them as additional demonstrations for
fine-tuning, combining the simplicity of supervised learning with the benefits
of closed-loop training. The key challenge is how to bias the rollouts towards
high-quality behavior such that the fine-tuning step improves the policy.  In
traffic simulation, Closest Among Top-K (CAT-K)~\cite{zhang2025closed}
instantiates this idea by selecting, at each step, the closest among a small set
of policy-proposed candidate actions to the ground-truth trajectory. On these
generated trajectories, CAT-K derives fine-tuning action targets using an
inverse dynamics model that chooses the action bringing the agent closest to
ground truth.

While effective for traffic modeling, CAT-K is poorly suited to modern E2E
policies because it assumes: (i) discrete actions; (ii) deterministic dynamics and known inverse dynamics;
(iii) a diverse pretrained policy where at least one action sample lies close to the ground-truth
trajectory at each time step; and (iv) that fresh on-policy trajectories can be generated
continuously during training. In E2E driving, none of these assumptions usually hold: policies
may output multi-token plans or continuous trajectories, as in the case of diffusion policies; 
the dynamics (including downstream controllers) are stochastic without closed-form inverse dynamics;
the action distribution is typically less diverse due to safety- and,
predictability-oriented training (unlike traffic agents that deliberately
promote diversity); and regenerating closed-loop rollouts at every optimization
step is prohibitively expensive.

\begin{figure}[t!]
  \includegraphics[width=1.0\linewidth]{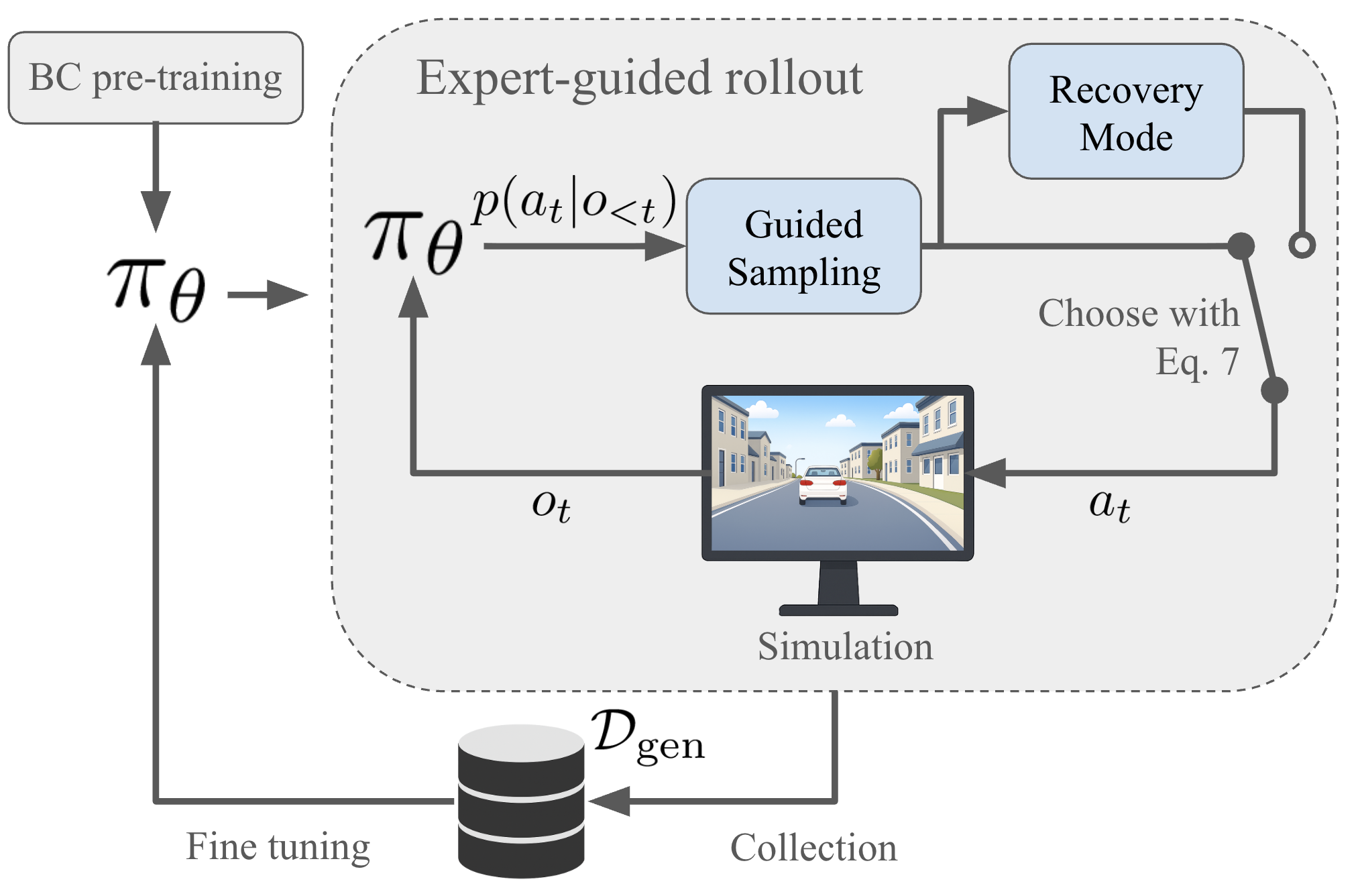}
  \caption{\textbf{RoaD method overview.} A pretrained policy generates rollouts in simulation via expert-guided sampling and an optional recovery mode. The collected trajectories form the CL-SFT dataset used to fine-tune the policy. The cycle can be repeated.}
  \label{fig:method_overview}
  \vspace{-1ex}
\end{figure}

In this work we introduce \textbf{Rollouts as Demonstrations (RoaD)}, a novel
CL-SFT approach that addresses all limitations above (\cref{fig:method_overview}). 
First, RoaD retains the expert-biased rollout principle, but removes the need for a recovery action
by treating the policy’s closed-loop rollouts directly as additional
demonstrations for SFT. Empirically we find that this strategy achieves performance on par with, or
better than CAT-K on large-scale traffic-simulation benchmarks. 
Second, we replace the Top-K selection with
sampling K action candidates so that RoaD can be applied to a more general class of policies. 
Third, when limited action diversity prevents naive
RoaD from being guided by the ground truth, we introduce a lightweight
recovery-mode policy output that enables following the ground-truth trajectory %
even when it is not close to any of the top-k most likely actions. Finally, to reduce
collection cost, we show that reusing rollout datasets across multiple
optimization steps results in only marginal performance degradation, greatly 
improving data efficiency. %

In experiments, we first validate RoaD for traffic simulation using the Waymo Open Simulation Agent Challenge (WOSAC). RoaD outperforms or matches CAT-K, even when it generates CL experiences only once with the base policy, and the performance improves further the more frequently the data is updated.
We then apply RoaD to an E2E driving task to fine-tune a VLM-based policy deployed in AlpaSim~\cite{nvidia2025alpasim}, an E2E AV simulator that reconstructs real-world 3D scenes using SOTA 3D Gaussian splatting, 3DGUT~\cite{wu20253dgut}. RoaD fine-tuning improves driving scores by 41\% and reduces collisions by 54\% over the base model in previously unseen scenarios. %

In summary, our conclusions are as follows.
\begin{itemize}
    \item We introduce a novel CL-SFT algorithm, RoaD, that removes restrictive assumptions made by prior work.
    \item We apply RoaD to traffic simulation and match or outperform the previous SOTA CL-SFT method.
    \item We apply RoaD to E2E driving and achieve substantial improvement in closed-loop driving metrics.
\end{itemize}

\section{Related work}
\label{sec:related_work}

\myparagraph{Closed-loop training in AV.} Classic closed-loop training methods such as DAgger \cite{ross2011reduction} and DART~\cite{laskey2017dart} iteratively collect states induced by the learner and query the expert to label them, thus adapting the training distribution to the learner’s roll-out distribution. 
However, in driving applications expert interventions are expensive, unsafe, or infeasible during interaction. Therefore, despite efforts on reducing expert burden in variants of the algorithm~\cite{ross2014reinforcement,sun2017deeplyaggrevated,zhang2017queryefficient,kahn2018self}, repeated expert relabeling  remains impractical in autonomous driving. 

Reinforcement learning presents a natural closed-loop training paradigm with a large body of literature in autonomous driving \cite{isele2018navigating,zhu2020safe,saxena2020driving,ma2021reinforcement}. From early works such as \cite{sallab2017deep} to more recent studies applying hierarchical RL \cite{chen2021interpretable}, curriculum-based RL \cite{codevilla2019exploring}, or model-based RL \cite{xu2020guided}, they face two major challenges. (1) reward design that captures the diverse and often conflicting requirements; (2) training stability, compute cost, and safety constraints for E2E driving systems. These practical issues often lead to slow convergence, brittle behavior, and poor transfer to rare scenarios when applying RL to AV. 

Inspired by recent work on CAT-K~\cite{zhang2025closed}, our method performs closed-loop training \textit{without} relying
on reinforcement rewards or on-demand expert relabeling. It derives training targets from expert demonstrations 
and performs supervised updates in the closed-loop setting, thereby achieving the stability and data-efficiency benefits of
supervised learning. %
Importantly, unlike CAT-K, our method does not require a discrete action space and deterministic dynamics, which extends its applicability to domains such as E2E driving.

\myparagraph{End-to-end driving.} 
Modern AV systems increasingly favor end-to-end designs over modular pipelines to reduce information loss and simplify training~\cite{karkus2022diffstack}. Work in this direction was pioneered by, among others, DAVE-2 \cite{bojarski2016end}. Subsequent approaches, such as ChauffeurNet \cite{bansal2019chauffeurnet}, UniAD \cite{hu2022planning}, and PARA-Drive \cite{weng2024paradrive} introduced intermediate BEV representations, and integrated perception, prediction, and planning within unified multi-task networks. %

In recent years, the emergent class of vision‐language‐action (VLA) models has adapted large vision–language models (VLMs) to driving, combining visual input, language instructions, and trajectory generation~\cite{jiang2025survey,zhou2024vlmsurvey,hwang2024emma,tian2024drivevlm,zhou2025autovla,wang2025alpamayo}. These models promise richer semantic understanding and human-aligned decision making. In particular, systems like EMMA~\cite{hwang2024emma} and Alpamayo-R1~\cite{wang2025alpamayo} demonstrate how VLM-based architectures can ingest camera imagery (and optionally language navigation cues) and output trajectories in a unified framework. While equipped with strong semantic understanding capabilities, these models are largely trained in open-loop, and thus remain prone to covariate shift.

\section{Background}
\label{sec:background}

Our goal is to finetune a pretrained driving policy in closed-loop 
to minimize the covariate shift between open-loop pre-training and closed-loop
deployment. To achieve this without access to a reward function (which is hard
to define for this task), we use closed‑loop supervised fine‑tuning (CL‑SFT).
In the following, we first formalize the problem and review CAT-K~\citep{zhang2025closed}, a recent CL‑SFT instantiation. %

\subsection{Problem formulation}
\label{sec:problem_formulation}

We are given a policy $\pi_\theta=p(a_t | o_{<t})$ that maps a history of observation inputs
$o_{<t}$ to action outputs $a_t$. The policy is pre-trained with behavior cloning
(BC), using a dataset of expert demonstrations $\mathcal{D} =
\{({o}^{E,i}_{0:T}, {a}^{E,i}_{0:T})\}_{i=1}^{|\mathcal{D}|}$.  Our
goal is to perform closed-loop finetuning of $\pi_\theta$ to minimize the
covariate shift between open-loop training and closed-loop deployment.  We
assume access to a stochastic simulator that generates a next observation given
 action, previous observation and some internal state $\mathcal{P}(o_{t+1} \mid
{o}_{t}, a_t; \cdot)$, but no access to an on-demand expert nor to a reward
function.

The contents of the observation $o_t$ are domain specific. In
traffic simulation, $o_t$ includes the positions, velocities, and orientations
of all nearby agents, together with a vectorized map (lane markings, wait lines, etc.);
in E2E driving, it includes the ego vehicle’s sensor inputs (e.g., multi-view
camera images) and estimated egomotion (e.g., pose, steering angle, velocity,
and acceleration). 
We denote by $s_t\in\mathbb{R}^{D_s}$ the pose of the controlled agent.
For notational simplicity, we assume
control of a single agent. Since RoaD operates per agent, extending to
multi-agent control, as in our traffic simulation experiments, is
straightforward.

Unlike prior work, we make no strong assumptions on the structure of $a_t$: it
may represent a state delta, as is common in
traffic simulation; a continuous control
signal, a waypoint, or a trajectory.  Single-step control inputs are
executed through forward dynamics, respecting
vehicle motion constraints, while waypoints and trajectories are tracked by
low-level controllers.  Predicting $T_{\mathrm{pred}}$-step trajectories
is common because such \emph{action chunking} encourages long‑horizon reasoning
and often improves accuracy with open-loop training. For
notational simplicity, we refer to all these outputs uniformly as $a_t$ and use
$s_{t+1}=f(s_t,a_t)$ to denote the agent state evolution over time.

Further, prior work assumed a policy with discrete modes to select top $K$
predictions, such as next-token-prediction (NTP) traffic simulation models that
encode actions in a single token.  In contrast, we only assume that $\pi_\theta$
can generate $K$ independent action samples, allowing for modern E2E driving
policies such as Transformers with simple Gaussian outputs, NTP models with
multiple tokens per action, or diffusion and flow-matching
policies~\cite{wang2025alpamayo}.

\subsection{Closed-loop supervised fine-tuning and CatK}
\label{sec:catk}

CL-SFT adapts a pretrained policy by behavior cloning on states encountered under its own closed-loop rollouts,
aligning the training distribution with deployment and mitigating covariate
shift. CAT-K \citep{zhang2025closed} provides a practical instantiation with two complementary
components: (i) recovery supervision, which defines action
targets that move the rollout back toward the expert trajectory at the visited on-policy rollout states; and
(ii) expert-proximal rollouts using top $K$ predictions, which bias action selection during rollouts to
remain close to the expert, so that the recovery supervision remains valid.
Intuitively, CAT-K learns “how to get back on track” while ensuring it never
drifts too far from the track in the first place.

Formally, the algorithm assumes a tokenized model or a distribution with discrete modes, which can be generally written as 
$\pi(a_t \mid o_{<t}) = \sum_{m=1}^{M} \pi(a \mid m)\, \pi(m \mid o_{<t})$,
where $M \in \mathbb{N}$ denotes the vocabulary size or number of modes, $\pi(m \mid o_{<t})$ the token prediction or mode-selection distribution, and $\pi(a \mid m)$ the action decoder or action distribution within each mode.
In each rollout step, the algorithm selects the top $K$ predictions, $\Xi_t=\topk{K}[\pi_\theta(a \mid o_{<t})]$, where $\topk{K}[\pi]$ represents finding the $K$ most likely tokens/modes under $\pi(m \mid o_{<t})$ and decoding/sampling the associated action.  
To bias rollouts toward the expert, the action ``closest'' to the expert is selected,
\begin{equation}
    a_t = \arg\min_{a\in\Xi_t} d\big(f(s_t,a), s^E_{t+1}\big),
    \label{eq:catk}
\end{equation}
where $s_t$ is the current agent state, $f$ are the deterministic dynamics, $s^E_{t+1}$ is the next expert state, and
$d(\cdot,\cdot)$ is a distance metric on states, e.g., a weighted $\ell_2$ over position, heading, and speed.
For each state, recovery actions are defined by projecting the expert continuation onto the action vocabulary, \TODO{define this with inverse dynamics instead (given we highlight its importance now earlier in the paper)}
\begin{equation}
    \hat{a}_t = \arg\min_{a\in\{1,\dots,|M|\}} d\big(f(s_t,a), s^E_{t+1}\big),
\end{equation}
and $\theta$ is updated with behavior cloning on the rollout states:
$\mathcal{L}_{\mathrm{BC}}(\theta) = - \frac{1}{NT} \sum_{t=0}^{T-1} \sum_{i=1}^{N} \log 
\pi_\theta(\hat{a}_t^{i} \mid o_{<t}),$
where $N$ represents the number of controlled agents. 

CAT-K is highly effective in achieving this goal for traffic
simulation, but is limited when applied to E2E driving, due to the assumptions of deterministic dynamics and known inverse dynamics to construct recovery action targets, and it's reliance on single-step, discrete policies to efficiently compute the top-K operator. %

\begin{table*}[ht!]
\setlength{\tabcolsep}{8pt}
\centering
\scalebox{\tablescaler}{
\begin{tabular}{lrccccc} 
\toprule
\begin{tabular}{@{}l@{}} \emph{WOSAC leaderboard, test split} \\ Method \end{tabular} 
& \begin{tabular}{@{}r@{}} \# model \\ params \end{tabular} 
& \begin{tabular}{@{}c@{}} RMM \\ $\uparrow$ \end{tabular} 
& \begin{tabular}{@{}c@{}} Kinematic \\ metrics $\uparrow$ \end{tabular} 
& \begin{tabular}{@{}c@{}} Interactive \\ metrics $\uparrow$ \end{tabular} 
& \begin{tabular}{@{}c@{}} Map-based \\ metrics $\uparrow$ \end{tabular}
& \begin{tabular}{@{}c@{}} min \\ ADE $\downarrow$ \end{tabular}  \\
\cmidrule(lr){1-2}\cmidrule(lr){3-3}\cmidrule(lr){4-7}
SMART-tiny RoaD (ours) & $7$ M
& \textbf{0.7847} & \textbf{0.4932} &  \textbf{0.8106} &  \textbf{0.9178}  & \textbf{1.3042} \\
SMART-tiny CAT-K~\cite{zhang2025closed} & $7$ M
& ${0.7846}$  & 0.4931 & \textbf{0.8106} & 0.9177  & 1.3065  \\
SMART-large~\cite{wu2025smart} & $102$ M
& $0.7614$ & $0.4786$ & $0.8066$ & $0.8648$  & $1.3728$  \\
SMART-tiny~\cite{wu2025smart} & $7$ M
& $0.7591$ & $0.4759$ & $0.8039$ & $0.8632$  & $1.4062$  \\
\bottomrule
\end{tabular}
}
\vspace{-1ex}
\caption{\textbf{WOSAC leaderboard \cite{wosac2024} for traffic simulation comparing CL-SFT approaches}. RMM stands for Realism Meta Metric, the key metric used for ranking.
RoaD fine-tuning significantly improves over the base model (SMART-tiny), it outperforms a much larger model from the same model family (SMART-large), and it is on par with the SOTA CL-SFT method, CAT-K.
}
\label{table:wosac_leaderboard_clsft}
\end{table*}

\begin{table*}[ht!]
\setlength{\tabcolsep}{8pt}
\centering
\scalebox{\tablescaler}{
\begin{tabular}{lrccccc} 
\toprule
\begin{tabular}{@{}l@{}} \emph{WOSAC local val. split} \\ Method \end{tabular} 
& \begin{tabular}{@{}r@{}} Data update \\ frequency \end{tabular} 
& \begin{tabular}{@{}c@{}} RMM \\ $\uparrow$ \end{tabular} 
& \begin{tabular}{@{}c@{}} Kinematic \\ metrics $\uparrow$ \end{tabular} 
& \begin{tabular}{@{}c@{}} Interactive \\ metrics $\uparrow$ \end{tabular} 
& \begin{tabular}{@{}c@{}} Map-based \\ metrics $\uparrow$ \end{tabular}
& \begin{tabular}{@{}c@{}} min \\ ADE $\downarrow$ \end{tabular}  \\
\cmidrule(lr){1-2}\cmidrule(lr){3-3}\cmidrule(lr){4-7}
RoaD fine-tuning & always
& \textbf{0.7673}  & \textbf{0.4871} & \textbf{0.8107} & ${0.8715}$  & ${1.3004}$  \\
RoaD fine-tuning & every 2 epochs
& $0.7669$  & $0.4865$ & $0.8098$ & \textbf{0.8720}  & \textbf{1.2893}  \\
RoaD fine-tuning & one-off
& $0.7664$  & $0.4865$ & $0.8093$ & $0.8712$  & $1.2983$  \\
SMART-tiny base model& --
& $0.7653$ & $0.4831$ & $0.8081$ & $0.8716$ & $1.3240$ \\
\midrule
CAT-K fine-tuning~\cite{zhang2025closed} & always
& $0.7616$  & $0.4583$ & $0.8105$ & \textbf{0.8720}  & $1.3105$  \\
SMART-tiny base model (from~\cite{zhang2025closed}) & --
& $0.7581$ & $0.4512$ & $0.8076$ & $0.8697$ & $1.3152$ \\
\bottomrule
\end{tabular}
}
\vspace{-1ex}
\caption{\textbf{Ablation of data collection frequency for traffic simulation, WOSAC 2\% validation split}. RoaD fine-tuning leads to significant improvement even when closed-loop data is only collected once, achieving similar levels of improvements over the base model as through CAT-K fine-tuning. The more frequently the data is updated the larger the performance gain. Note that results for CAT-K were taken from~\cite{zhang2025closed}, where likely a different SMART-tiny checkpoint was used as a base model.
}
\label{table:wosac_ablation}
\vspace{-2ex}
\end{table*}

\section{Method}
\label{sec:method}

\begin{algorithm}[t]
\footnotesize
\caption{RoaD}
\label{alg:road}
\begin{algorithmic}[1]
    \State \textbf{Input}: policy $\pi_\theta$, dataset $\mathcal{D}$, candidate action set size $K$, number of rollouts $N_\mathrm{roll}$, number of training steps $N_\mathrm{train}$, recovery parameters $(\delta_{\mathrm{rec}}, N_{\text{rec}})$
        \State Initialize dataset $\mathcal{D}_{\text{gen}}=\{\}$
        \For{$j=0,\dots,N_\mathrm{roll}$} 
            \State Start simulation with scenario $s^E_{0:T} \sim \mathcal{D}$ 
            \For{$t=1,\dots,T-1$}
                \State Sample candidates (\cref{eq:samplek})
                \State Choose closest to expert (\cref{eq:gen_distance})
                \If{trigger (\cref{eq:recovery_trigger})}
                    \State Use recovery mode output $a_t \gets a'_t$ (\cref{eq:recovery_blend})
                \EndIf
                \State Step simulator $o_{t+1} \sim \mathcal{P}(o_{t+1} | a_t, o_t; \cdot)$
            \EndFor
            \State Add rollout to dataset $\mathcal{D}_{\text{gen}} \gets (o_{0:T}, a_{0:T})$
        \EndFor
        \For{$j=0,\dots,N_\mathrm{train}$} 
            \State Update $\theta$ with $(o_{t}, a_{t})\sim\mathcal{D}_{\text{gen}}$ and the RoaD loss (\cref{eq:road_loss})
        \EndFor
\end{algorithmic}
\end{algorithm}

Our goal is a CL-SFT recipe that works with modern E2E driving
policies and reduces the covariate shift between open-loop training and
closed-loop deployment without requiring a reward function.
Our proposed method, rollouts as demonstrations (RoaD), keeps CAT-K’s bias-toward-expert
idea but removes its main constraints while remaining simple and data-efficient.

\subsection{Rollouts as demonstrations (RoaD)}

The key idea of RoaD is to treat the policy’s own expert‑guided, closed‑loop
rollouts as additional supervision for fine‑tuning.
Formally, let $\mathcal{R}_{{s}^E_{0:T}}^{\mathcal{P}}[\pi_\theta]$ denote the expert‑guided
rollout operator for $\pi_\theta$ given the simulator $\mathcal{P}$ and expert (GT) trajectory
${s}^E_{0:T}$.
We accumulate generated rollouts in a dataset,
$$\mathcal{D}_{\text{gen}} = \Big\{ (o_{0:T}, a_{0:T}) \,\Big|\, o_{0:T}, a_{0:T} \sim
\mathcal{R}_{s^E_{0:T}}^{\mathcal{P}}[\pi_\theta], \,\,
{s}^E_{0:T} \sim \mathcal{D}\Big\}$$
and fine‑tune the policy by behavior cloning:
\begin{equation}
\mathcal{L}_{\mathrm{RoaD}}(\theta) = - \sum_{(o^i, a^i)\in \mathcal{D}_{\text{gen}}} \sum_{t} \sum_{i=1}^{N} \log \pi_\theta(a_t^i \mid o_{<t}).
\label{eq:road_loss}
\end{equation}

The expert guidance is designed to produce trajectories that are
simultaneously near on‑policy (i.e. sampled from $\pi_\theta$), but also higher‑quality than
unassisted rollouts. Because this data is collected on-policy, it
covers states the policy is likely to encounter,
reducing covariate shift between training and deployment.
In practice, $\mathcal{R}_{{s}^E_{0:T}}[\pi_\theta]$ can be implemented by
biasing the policy’s output toward the expert continuation, for example using
Top‑$K$ selection (\cref{eq:catk}) or Sample‑$K$ (see \cref{sec:sampleK}).

Crucially, compared to CAT-K, RoaD does not require the construction of target
recovery actions which are challenging to construct under stochastic or non-invertible dynamics, and are often low-quality for policies that output future trajectories rather than single-step actions.
Instead, it uses the future trajectory itself as the target.
In the following, we discuss three further modifications which makes
RoaD applicable to E2E driving. %

\subsection{Sample-$K$ expert-guided rollouts}
\label{sec:sampleK}

To preserve expert guidance without discrete top-$K$ enumeration, we draw $K$
action candidates from the current policy distribution (e.g., trajectory samples or
diffusion/flow-matching draws), 
\begin{equation}
    \Xi_t = \{a_t^{(k)}\}_{k=1}^K \sim \text{i.i.d. } \pi_\theta(\cdot\mid o_{<t})
\label{eq:samplek}
\end{equation}
and select the candidate closest to the expert continuation under a generalized distance metric (\cref{eq:generalized_distance}):
\begin{equation}
    a_t = \arg\min_{a\in\Xi_t} d^{\mathrm{g}}\!\big(a, \, s^E_{t:T}\big).
    \label{eq:gen_distance}
\end{equation}
This Sample-K relaxation maintains the “closest-to-expert” bias while accommodating continuous policy outputs
and large vocabularies. 

When $a_t$ represent trajectories, the distance function $d(\cdot,\cdot)$ can be implemented as a trajectory-level (generalized) distance between predicted trajectories and the future expert trajectory. A concrete choice is a weighted step-wise distance over a comparison horizon $H_t$,
\begin{equation}
d^{\mathrm{g}}\!\big(a_t,\, {s}^E_{t:T}\big) = \sum_{k=1}^{H_t} w_k\; d\big(\tilde{s}_{t+k}(a_t),\, {s}^E_{t+k}\big),
\label{eq:generalized_distance}
\end{equation}
where $\tilde{s}_{t+k}(a_t)$ denotes the predicted state at step $t+k$ implied by action $a_t$,
and $w_k\!\ge\!0$ are arbitrary weights. 

\subsection{Recovery-mode policy output}
E2E driving policies often exhibit limited action diversity as they are trained
to drive safely and predictably, preventing naive sampling from reliably
producing a candidate near the expert. To address this, we introduce an optional
recovery-mode policy output that nudges the policy toward the expert when all
sampled actions are too far from the expert.

Concretely, when the chosen action $a_t$ is a trajectory, we linearly
interpolate between $a_t$ and the expert continuation, acting as guidance
rather than a discrete override. Let the prediction horizon be $F$. We reuse the
notation $\tilde{s}_{t+k}(a_t)$ from \cref{eq:generalized_distance} to denote the
predicted state at step $t+k$ implied by $a_t$.
Recovery is triggered when the generalized distance to the expert exceeds a
threshold:
\begin{equation}
    d^{\mathrm{g}}\!\big(a_t,\, s^E_{t:T}\big) > \delta_{\mathrm{rec}}.
    \label{eq:recovery_trigger}
\end{equation}
Upon triggering, we define a weight vector $\lambda\in[0,1]^F$ (e.g., a linear
schedule $\lambda_k=\min(1, k/N_{\text{rec}})$) and blend the trajectories as
\begin{equation}
    \tilde{s}_{t+k}(a'_t) = (1-\lambda_k)\, \tilde{s}_{t+k}(a_t) + \lambda_k\, {s}^E_{t+k},\quad k=1,\dots,F,
    \label{eq:recovery_blend}
\end{equation}
which defines the recovery trajectory $a'_t$. The weight vector $\lambda$
subsumes all parameters of the schedule; in practice we use a simple linear
ramp over the first $N_{\text{rec}}$ steps.

\begin{table*}[ht!]
\setlength{\tabcolsep}{8pt}
\centering
\scalebox{\tablescaler}{
\begin{tabular}{lrcccc} 
\toprule
\begin{tabular}{@{}l@{}} \emph{AV NuRec dataset} \\ Method \end{tabular} 
& \begin{tabular}{@{}r@{}}  \end{tabular} 
& \begin{tabular}{@{}c@{}} Driving \\ score $\uparrow$ \end{tabular} 
& \begin{tabular}{@{}c@{}} Collision \\  rate $\downarrow$ \end{tabular} 
& \begin{tabular}{@{}c@{}} Offroad \\ rate $\downarrow$ \end{tabular} 
& \begin{tabular}{@{}c@{}} Distance \\ traveled (m) $\uparrow$ \end{tabular}
\\
\cmidrule(lr){1-2}\cmidrule(lr){3-3}\cmidrule(lr){4-6}
Fine-tuning with RoaD (ours) & 
& \textbf{0.6300\std{0.0090}}   & \textbf{0.0239\std{0.0000}} & \textbf{0.2098\std{0.0029}} & 147.2\std{0.54} \\
Fine-tuning with re-rendered expert trajectories & 
& 0.4985\std{0.0046}   & 0.0464\std{0.0051} & 0.2583\std{0.0044} & \textbf{151.9\std{0.36}} \\
Continued large-scale training with BC & 
& 0.4215\std{0.0092}  & 0.0627\std{0.0033} & 0.2783\std{0.0039} & 143.7\std{0.76} \\
Base model pre-trained with BC & 
& 0.4443\std{0.0210}   & 0.0525\std{0.0051} & 0.2833\std{0.0084} & 149.0\std{1.08} \\
\bottomrule
\end{tabular}
}
\vspace{-1ex}
\caption{\textbf{End-to-end simulation results over the AV NuRec dataset.} RoaD fine-tuning significantly increases the driving score, and it outperforms both continued open-loop training with real data, as well as fine-tuning with re-rendered expert trajectories.}
\label{table:alpasim_results}
\end{table*}

\begin{table*}[ht!]
\setlength{\tabcolsep}{8pt}
\centering
\scalebox{\tablescaler}{
\begin{tabular}{lrcccc} 
\toprule
\begin{tabular}{@{}l@{}} \emph{AV NuRec dataset} \\ Method \end{tabular} 
& \begin{tabular}{@{}r@{}}  \end{tabular} 
& \begin{tabular}{@{}c@{}} Driving \\ score $\uparrow$ \end{tabular} 
& \begin{tabular}{@{}c@{}} Collision \\  rate $\downarrow$ \end{tabular} 
& \begin{tabular}{@{}c@{}} Offroad \\ rate $\downarrow$ \end{tabular} 
& \begin{tabular}{@{}c@{}} Distance \\ traveled (m) $\uparrow$ \end{tabular}
\\
\cmidrule(lr){1-2}\cmidrule(lr){3-3}\cmidrule(lr){4-6}
RoaD & 
& \textbf{0.6300\std{0.0090}}   & \textbf{0.0239\std{0.0000}} & \textbf{0.2098\std{0.0029}} & 147.2\std{0.54} \\
RoaD (no expert guidance) &   %
& 0.4847\std{0.0027} & 0.0576\std{0.0022} & 0.2543\std{0.0011}  & 151.2\std{0.24}\\
RoaD (no recovery) & 
& 0.5030\std{0.0107}  & 0.0518\std{0.0044} & 0.2493\std{0.0013} & \textbf{151.4\std{0.42}} \\
\midrule
RoaD (1 rollout) & 
 & 0.5898\std{0.0108} & 0.0341\std{0.0006} & 0.2091\std{0.0045} & 143.4\std{0.29} \\
RoaD (3 rollouts, default) & 
& 0.6300\std{0.0090}   & \textbf{0.0239\std{0.0000}} & 0.2098\std{0.0029} & 147.2\std{0.54} \\
RoaD (9 rollouts) & 
& \textbf{0.6317\std{0.0070}} & 0.0264\std{0.0027} & \textbf{0.2083\std{0.0054}} & \textbf{148.3\std{0.41}} \\
 \midrule
RoaD (fine-tune once, default) & 
& 0.6300\std{0.0090}   & \textbf{0.0239\std{0.0000}} & 0.2098\std{0.0029} & 147.2\std{0.54} \\
RoaD (fine-tune twice) & 
 & \textbf{0.6613\std{0.0130}} & 0.0420\std{0.0035} & \textbf{0.1967\std{0.0043}} & \textbf{157.8\std{0.37}} \\
\midrule
RoaD (1k steps) & 
 & 0.6171\std{0.0124} & 0.0246\std{0.0035} & 0.2152\std{0.0011} & 148.0\std{0.66}\\
RoaD (4.2k steps, default) & 
& \textbf{0.6300\std{0.0090}}   & \textbf{0.0239\std{0.0000}} & 0.2098\std{0.0029} & 147.2\std{0.54} \\
RoaD (10k steps) & 
 & 0.6095\std{0.0245} & 0.0424\std{0.0029} & \textbf{0.2043\std{0.0066}} & \textbf{150.2\std{0.15}} \\
\midrule
RoaD (K=16) & 
 & 0.5789\std{0.0039} & 0.0322\std{0.0023} & 0.2207\std{0.0029} & 146.4\std{0.18} \\
RoaD (K=32) & 
 & 0.5898\std{0.0060} & 0.0290\std{0.0006} & 0.2196\std{0.0033} & 146.6\std{0.16} \\
RoaD (K=64, default) & 
& 0.6300\std{0.0090}   & \textbf{0.0239\std{0.0000}} & 0.2098\std{0.0029} & 147.2\std{0.54} \\
RoaD (K=128) & 
& \textbf{0.6396\std{0.0119}}  & 0.0304\std{0.0039} & \textbf{0.2047\std{0.0045}} & \textbf{150.4\std{0.10}} \\
\midrule
Base model & 
& 0.4443\std{0.0210}   & 0.0525\std{0.0051} & 0.2833\std{0.0084} & 149.0\std{1.08} \\
\bottomrule
\end{tabular}
}
\vspace{-1ex}
\caption{\textbf{Ablation study for E2E driving.} The setting is identical to the main experiments apart from the ablated property. \emph{steps} refer to the number of optimization steps for fine-tuning. K denotes the number of trajectory samples for expert-guided rollouts. 
Results indicate that both expert guidance and recovery mode are important in the algorithm; and performance gains are observed over a wide range of hyperparameters.}
\label{table:alpasim_ablation}
\end{table*}

\subsection{CL-SFT with off-policy data}

CAT-K regenerates rollouts at each gradient step, which is feasible in BEV traffic
simulation, but prohibitive for E2E driving due to the high cost of rendering
sensor inputs. To reduce
this collection cost, we evaluate reusing the same rollout dataset across
multiple optimization steps, similar to a replay buffer in off-policy RL,
including the extreme case of generating only a single dataset at the start of
fine-tuning. Empirically, we find that rollout data reuse incurs only small degradation,
making RoaD practical when high‑fidelity rollouts are expensive to obtain.

\section{Experiments}
We validate our method for traffic simulation on the WOSAC benchmark~\cite{montali2023waymo}, and for E2E driving using the AlpaSim simulator~\cite{nvidia2025alpasim} and the NVIDIA Physical AI - AV NuRec Dataset~\cite{nvidia2025nurecavdata}.

\subsection{Traffic simulation}

We first validate RoaD for traffic simulation using the WOSAC benchmark.
Note that our primary goal here is not to outperform CAT-K, but to show that the
simplified RoaD approach can achieve comparable performance to CAT-K while also
being applicable to E2E driving due to fewer restrictive assumptions.

\subsubsection{Experimental setup}
We follow the experimental setup of \cite{zhang2025closed} and use RoaD to
fine-tune the SMART-tiny model on the WOMD dataset. The model receives 1 second
of trajectory history for all agents, it outputs delta x-y actions, and at test
time it is rolled out for 8 seconds with 0.1s time steps. %
Note that for this experiment we do not use the recovery mode as traffic models
are naturally diverse enough.

\myparagraph{Metrics.}
Evaluation follows the WOSAC protocol. For each scenario, we generate 32 rollouts for all agents in the scene and compare the resulting joint behavior distribution to human driven trajectories. 
We report the following metrics which, excluding minADE, measure the distributional similarity between the policy and the data.
The principal metric on the leaderboard is \textbf{Realism Meta Metric (RMM)}, which combines three distributional metrics: \textbf{kinematic metrics}, e.g. velocities and accelerations; \textbf{interactive metrics}, e.g., collisions; and \textbf{map-based metrics}, e.g. off-road driving. For the exact definition of the metrics we refer to \citep{montali2024waymo}. 
Additionally we also report \textbf{minADE}, i.e., minimum Average Displacement Error, a widely used metric for trajectory prediction.

\subsubsection{Results}
\myparagraph{Main results on WOSAC.} 
\cref{table:wosac_leaderboard_clsft} provides results on the public WOSAC leaderboard. RoaD fine-tuning significantly improves over the base model (SMART-tiny), it outperforms a much larger model from the same model family (SMART-large), and it is on-par with the SOTA CL-SFT method, CAT-K~\cite{zhang2025closed}. We note that multiple works on the leaderboard, concurrently developed with ours, such as SMART-R1~\cite{pei2025advancing}  achieve higher RMM using a combination of CL-SFT with CAT-K, and RL fine-tuning.  However, these approaches require highly specialized rewards derived from the WOSAC evaluation metrics, and a large number of environment interactions, making them unsuitable for E2E driving. A snapshot of the complete leaderboard at the time of submission is included in the Appendix. 

\myparagraph{Re-using CL experience.} To assess the effect of reusing previously
generated CL-SFT data, we ablate the data refresh frequency in
\cref{table:wosac_ablation}, evaluating locally on 2\% of the WOMD validation
set following \cite{zhang2025closed}. As expected, more frequent refreshes yield
higher performance, though the incremental gains are modest. Importantly, even
when closed-loop data is generated only once at the start of fine-tuning, RoaD
already delivers a substantial improvement. Given the high cost of data
rendering, this motivates our default E2E setup (\cref{sec:e2e_driving}) of generating CL-SFT data only once. For completeness, we also ablate repeated data generation and observe
additional, albeit smaller, improvements in the E2E experiment.

\begin{table}[t!]
\setlength{\tabcolsep}{4pt}
\centering
\scalebox{\tablescaler}{
\begin{tabular}{lrcc} 
\toprule
\begin{tabular}{@{}l@{}} \emph{Local scene set} \\ Method \end{tabular} 
& \begin{tabular}{@{}r@{}}  \end{tabular} 
& \begin{tabular}{@{}c@{}} 3D-GS  \\ Driving score $\uparrow$ \end{tabular} 
& \begin{tabular}{@{}c@{}}  NeRF %
  \\ Driving score $\uparrow$ \end{tabular} 
\\
\cmidrule(lr){1-2}\cmidrule(lr){3-4}
RoaD (ours) & 
& \textbf{0.75\std{0.23}} &  \textbf{0.58\std{0.09}} \\
Re-rendered expert trajectories & 
& 0.42\std{0.07} &   0.35\std{0.05}  \\
Base model & 
& 0.28\std{0.05} & 0.33\std{0.04} \\
\bottomrule
\end{tabular}
}
\vspace{-1ex}
\caption{\textbf{Sim2sim transfer results.} Policies are fine-tuned with 3DGS generated data, and evaluated in previously unseen 75 scenarios reconstructed either as 3DGS (default setting) or as a NeRF (sim2sim transfer). As expected, performance reduces when transferring fine-tuned policies to a new simulation environment, but fine-tuning with RoaD improves over the base model even in the transfer setting.}
\vspace{-1ex}
\label{table:alpasim_sim2sim}
\end{table}

\subsection{End-to-end driving}
\label{sec:e2e_driving}

Our main result is that CL-SFT with RoaD can significantly improve closed-loop performance in E2E driving.

\subsubsection{Experimental setup.}
\myparagraph{End-to-end VLA policy}
We employ a VLA-based policy structured similar to~\cite{wang2025alpamayo}. The policy takes in 1.6s ego motion history, and a sequence of timestamped images from two onboard cameras (front facing wide-angle and tele camera), and generates 6.4s trajectory sample output, which is then tracked by a downstream controller when executed in closed-loop. The policy is trained with a large-scale dataset comprising of 20,000 hours of human driving data from 25 countries, covering a variety of scenarios including highway and urban driving, weather conditions, day and night times. A 1700+ hour subset of this dataset is publicly available \citep{nvidia2025avdata}.

\begin{figure*}[ht!]
\setlength{\tabcolsep}{4pt}
\centering
\resizebox{\linewidth}{!}{%
\begin{tabular}{r c c c c} 

& $t=0s$ 
& $t=3s$
& $t=5s$ 
& $t=6s$ \\

\cmidrule(lr){2-5} 

\textbf{Pre-trained VLA policy} &
\includegraphics[width=0.3\textwidth, valign=c]{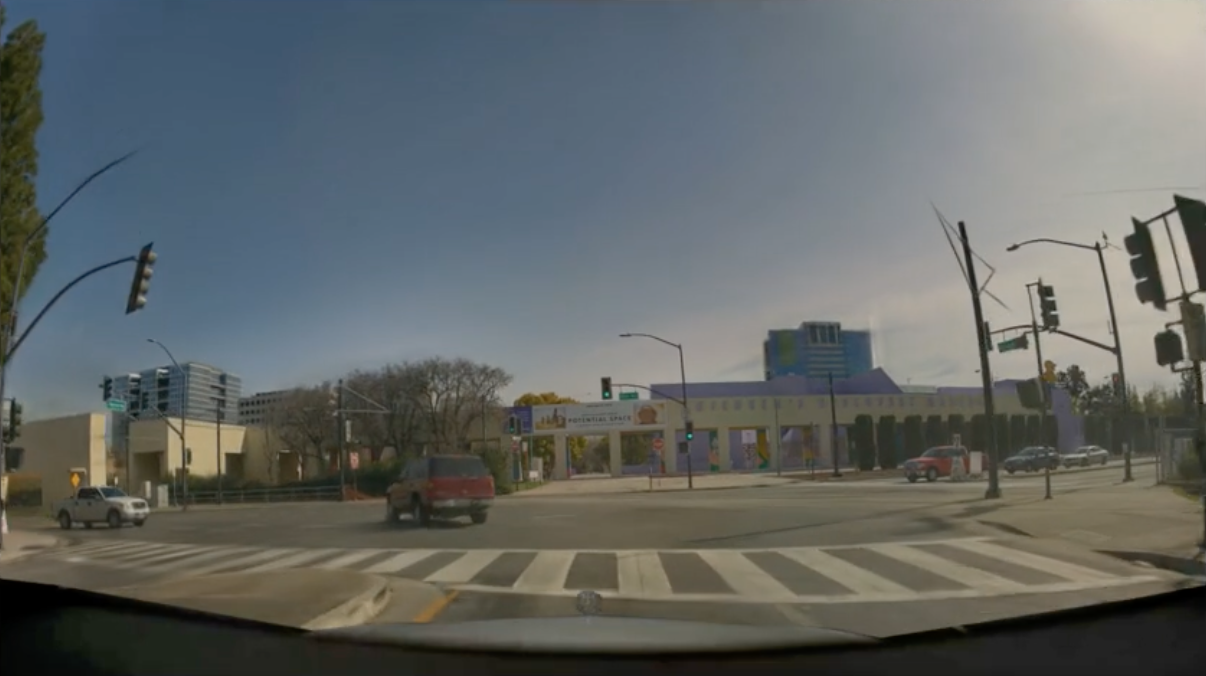} &
\includegraphics[width=0.3\textwidth, valign=c]{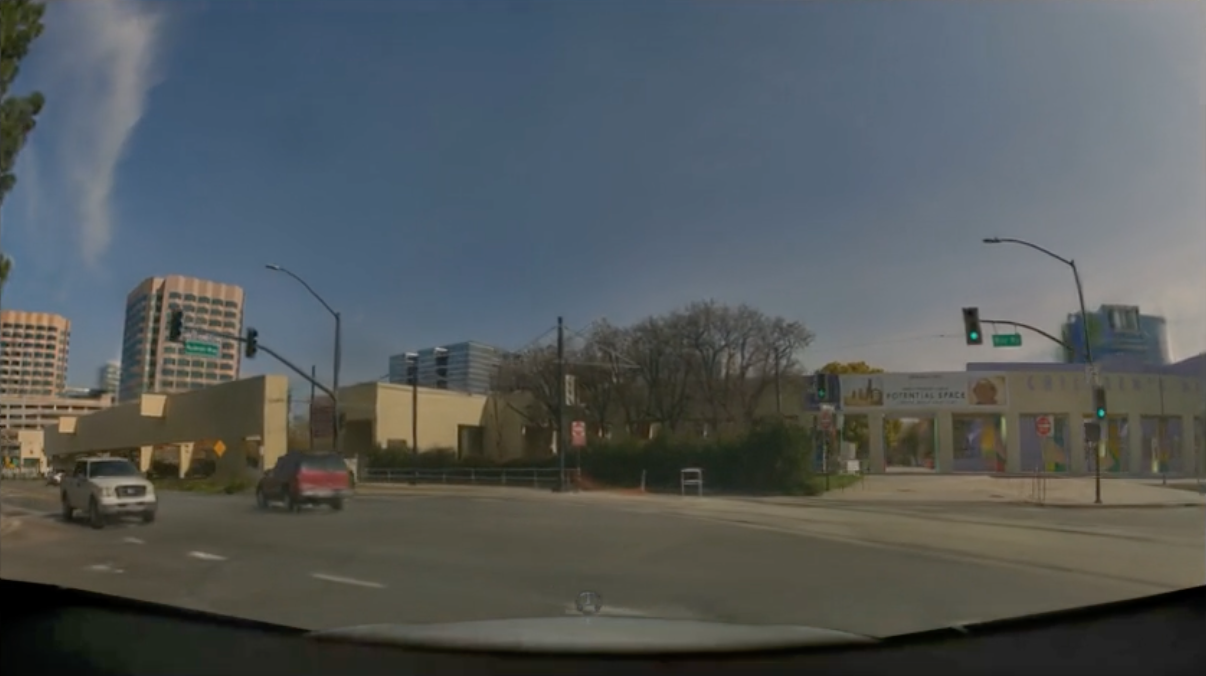} &
\includegraphics[width=0.3\textwidth, valign=c]{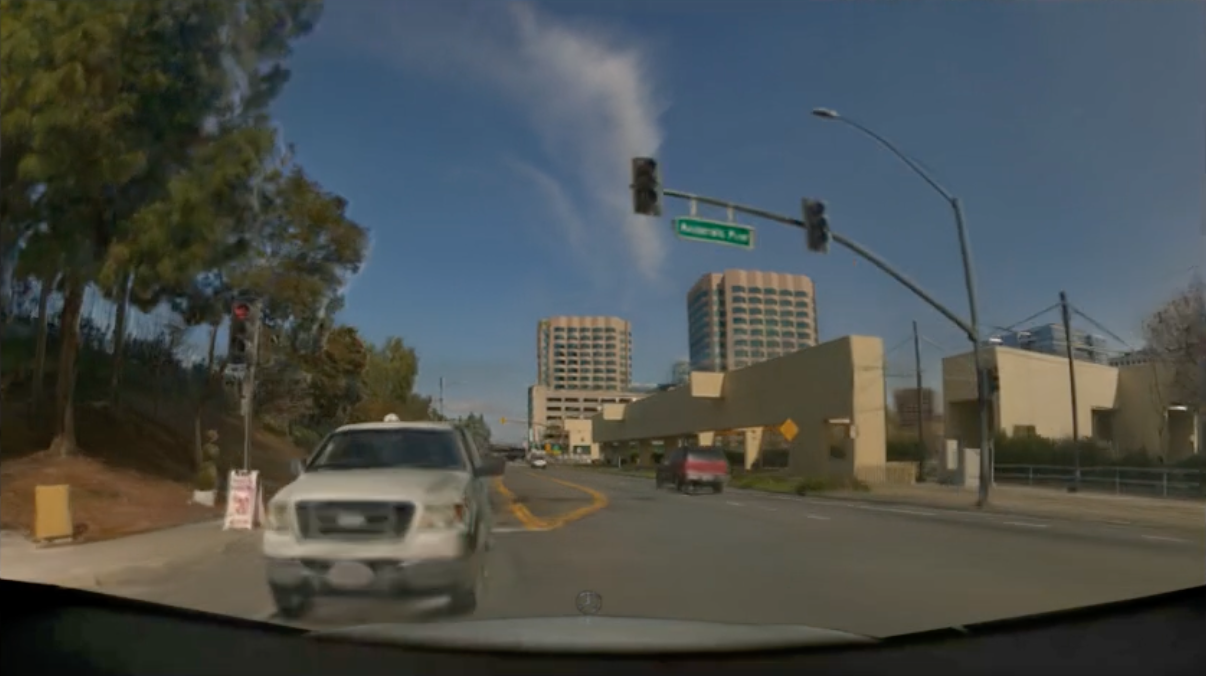} &
\includegraphics[width=0.3\textwidth, valign=c]{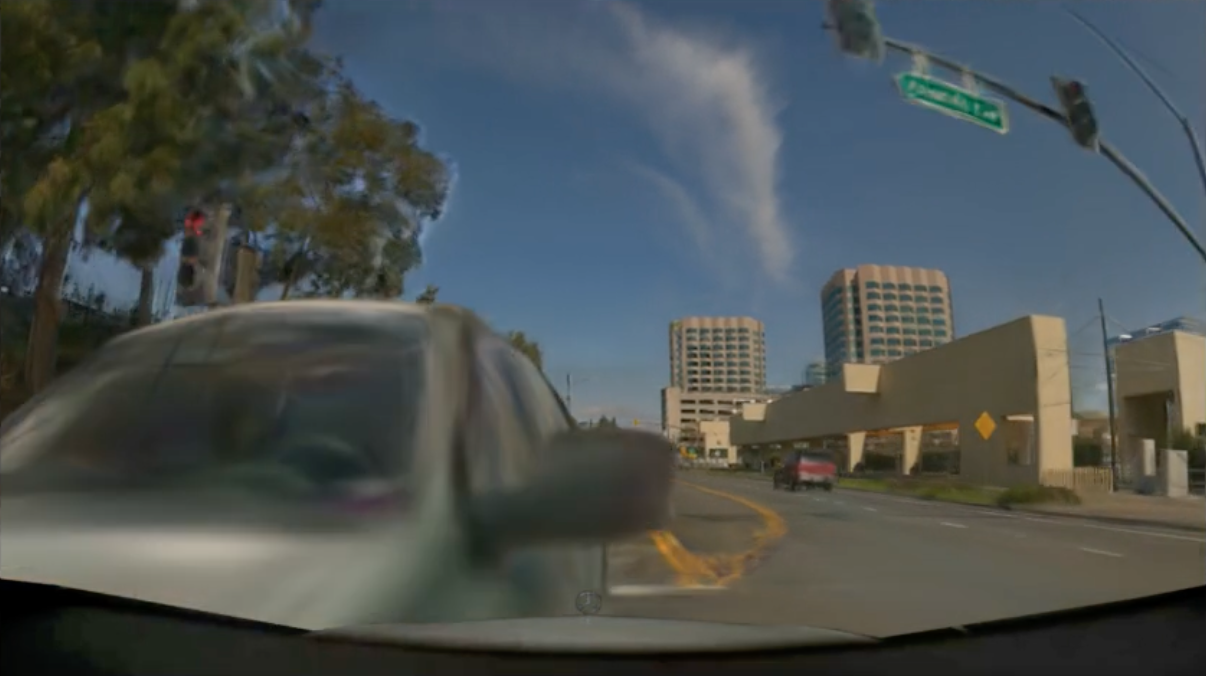} \\

\addlinespace[1.5ex] 

\textbf{RoaD fine-tuned VLA policy} &
\includegraphics[width=0.3\textwidth, valign=c]{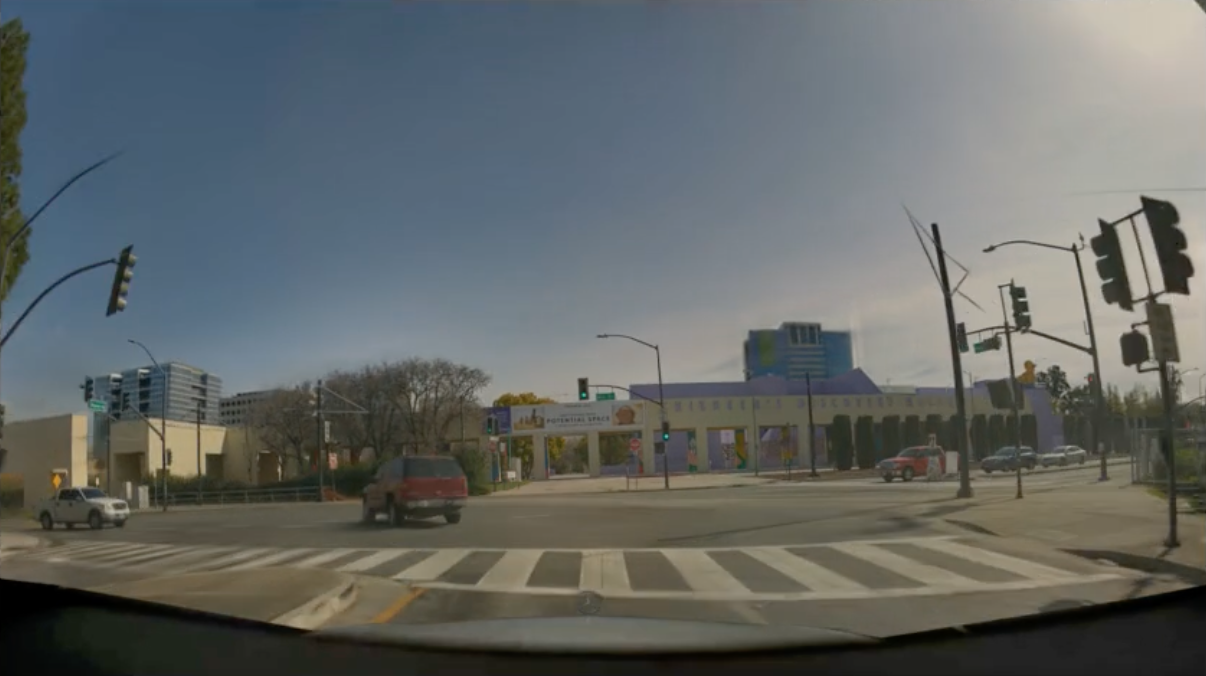} &
\includegraphics[width=0.3\textwidth, valign=c]{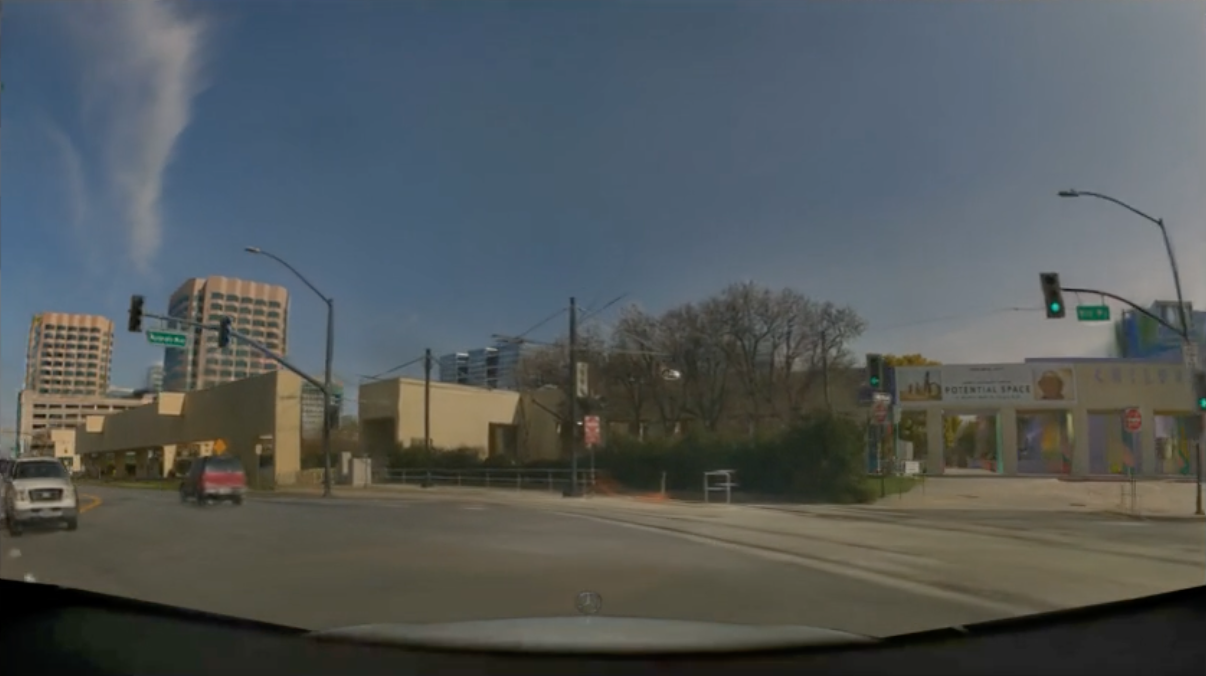} &
\includegraphics[width=0.3\textwidth, valign=c]{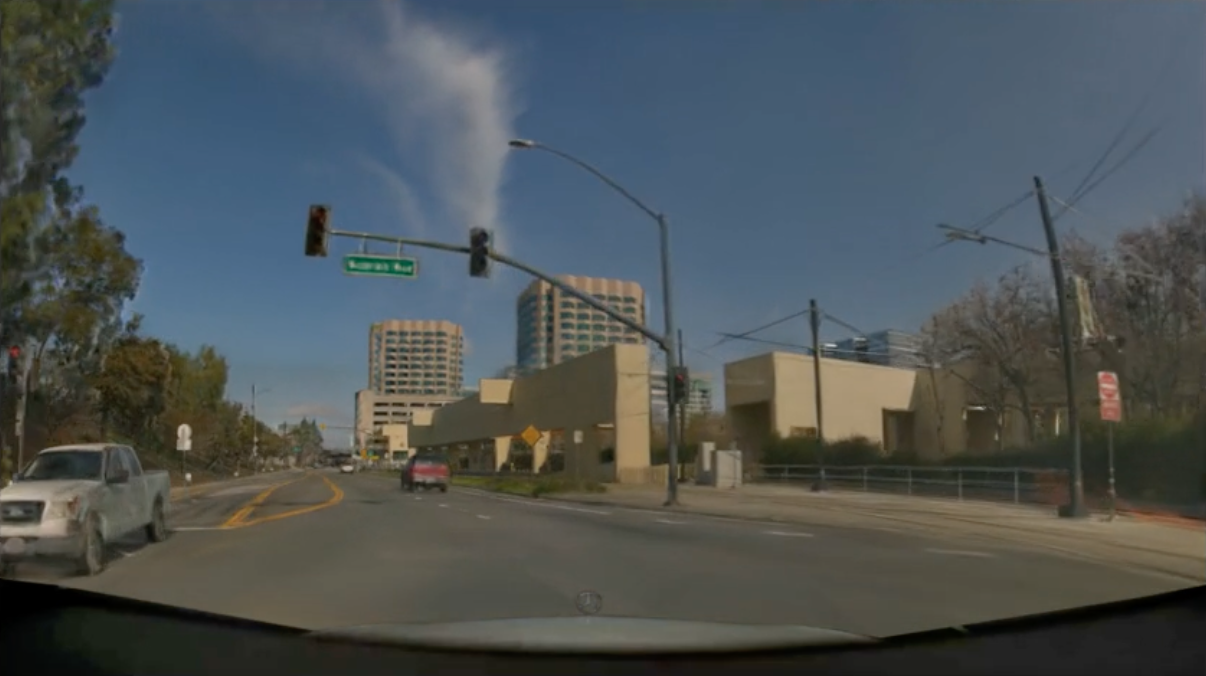} &
\includegraphics[width=0.3\textwidth, valign=c]{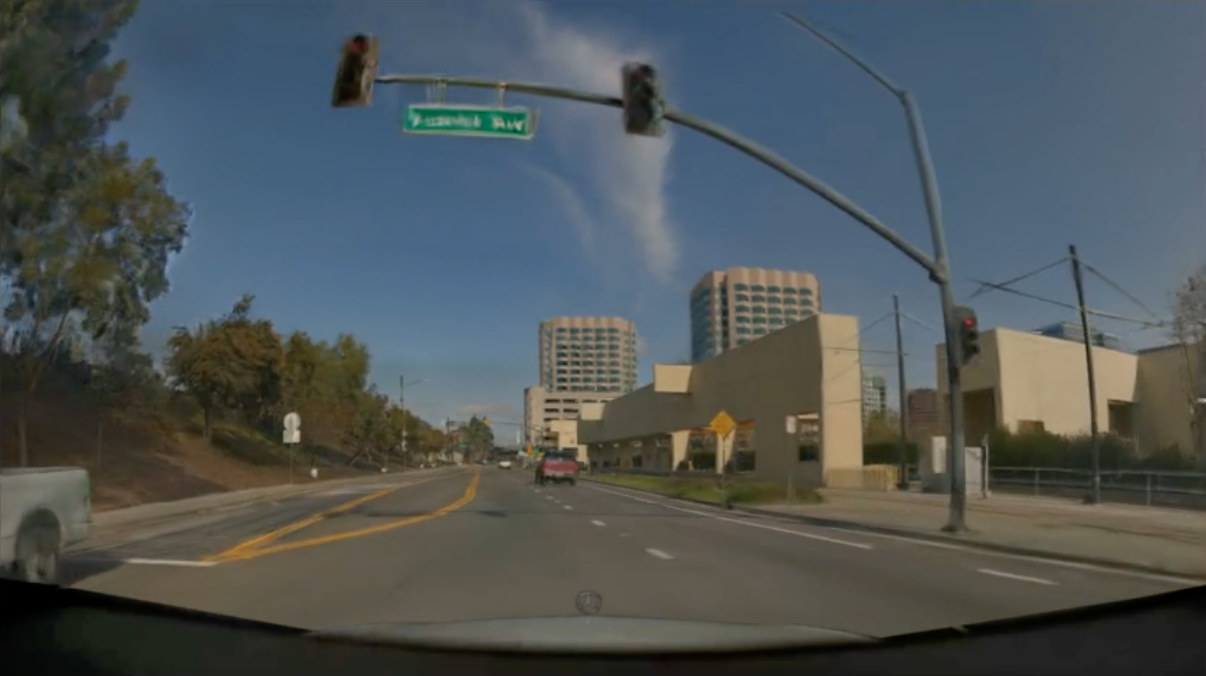} \\

\end{tabular}}
\caption{\textbf{Qualitative comparison of policy rollouts in our E2E simulator}. \textbf{Top row:} before fine-tuning, the policy navigates this intersection poorly, ends up in a wrong lane and fails to avoid a collision with a stationary vechicle. \textbf{Bottom row:} after fine-tuning with RoaD, the policy handles the intersection correctly and avoids any collision.}
\label{fig:qualitative_comparison}
\vspace{-2ex}
\end{figure*}

\myparagraph{Simulation environment.}
For E2E driving experiments, we employ AlpaSim~\cite{nvidia2025alpasim}, a closed-loop simulator built on SOTA neural scene reconstruction~\citep{wu20253dgut}. The system reconstructs real-world driving logs as temporal 3D Gaussian Splatting (3D-GS) scenes and renders novel camera views when the ego vehicle diverges from the recorded path. We employ custom controllers that track predicted trajectories with separate lateral and longitudinal control, using a 200 ms control delay and ego-motion noise. The vehicle dynamics is governed by a dynamically extended bicycle model.
The controller, control delay and dynamics model are designed to imitate real-word driving as closely as possible.
All other traffic participants, including vehicles and pedestrians, replay their logged trajectories.
Qualitative examples from AlpaSim~\cite{nvidia2025alpasim} are shown in \cref{fig:qualitative_comparison}. %

\myparagraph{Fine-tuning.}
To fine-tune the VLA policy we generate 20s long simulated CL data using 8251 3D-GS scenes reconstructed from real-world driving logs, 3x rollouts per scene by default.
We fine-tune for 4.2k steps (approximately one epoch of non-overlapping trajectory data) with frozen encoders to mitigate overfitting to visual artifacts.

\begin{figure}[t!]
  \centering
   \vspace{-2ex}
   \includegraphics[width=0.75\linewidth]{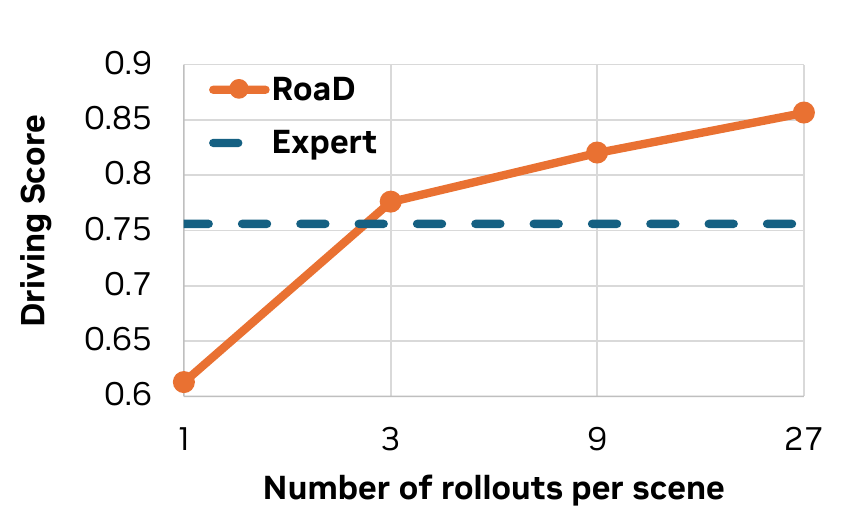}
   \vspace{-1.5ex}
   \caption{\textbf{The impact of generating multiple rollouts per scene.} In this experiment the same set of AV NuRec scenes are used for training and evaluation. RoaD performance improves monotonically as more rollouts are added to its SFT dataset (orange), while fine-tuning with resimulated expert demonstrations cannot make use of multiple rollouts.}
   \label{fig:rollout_vs_score}
   \vspace{-2ex}
\end{figure}

\myparagraph{Metrics}
To evaluate models, we use 920 openly accessible challenging 3D-GS scenes from the NVIDIA Physical AI - AV NuRec Dataset~\cite{nvidia2025nurecavdata}, and generate 3 rollouts per scenes. In \cref{table:alpasim_results,table:alpasim_ablation,table:alpasim_sim2sim}, mean values are computed over all scenes and rollouts, standard deviations are computed by taking the mean across scenes and computing the standard deviation across rollouts, estimating the evaluation uncertainty.
\canskip{The standard deviation over the full RoaD training, including data generation, fine-tuning, and evaluation with three rollouts per scene, is too expensive to perform for every model. For the driving score of our main result in \cref{table:alpasim_results} we found it to be $\pm 0.0057$.}

We report the following metrics. 
The \textbf{driving score} measures the average distance traveled (in kilometers) between incident events, where an incident corresponds to either off-road driving or a collision. 
The \textbf{collision rate} denotes the proportion of scenarios in which the ego vehicle is involved in a close encounter or collision for which it is deemed responsible, i.e., excluding rear-end and side contacts. 
The \textbf{off-road rate} captures the fraction of scenarios where the ego vehicle leaves the drivable area; this value appears relatively high because in the AV NuRec Dataset only the region bounded by lane markings is considered drivable. 
Finally, the \textbf{distance traveled} denotes the distance traveled by the ego vehicle in meters.

Each simulation terminates after the first incident. Evaluation in reconstructed scenes is inherently sensitive to rendering artifacts, particularly when the ego vehicle diverges from the logged path. To reduce the impact of such artifacts, we exclude any events where the ego deviates by more than 4\,m from the original trajectory. Nonetheless, a portion of recorded incidents remain attributable to visual artifacts or imperfect scene reconstructions.

\subsubsection{Results}

\myparagraph{Main results.} The main results for E2E driving are reported in \cref{table:alpasim_results}. RoaD fine-tuning increases the driving score in previously unseen scenarios by 41\% and reduces collisions by 54\%. 
RoAD outperform fine-tuning with expert demonstrations re-rendered in the same simulation environment, indicating that the performance gains of RoaD are not only from adjusting to simulation artifacts. 
RoaD also outperforms continued large-scale open-loop training using real-world driving data, indicating that the performance gains are not simply due to further training steps.
\cref{fig:qualitative_comparison} shows a qualitative example: while the base policy encounters a collision, after fine-tuning with RoaD, the policy handles the intersection correctly without collision. 

\myparagraph{Ablation studies.} Ablation results in \cref{table:alpasim_ablation} indicate that both expert guidance during rollouts and recovery-mode policy outputs are important for best performance in E2E driving.
Furthermore, RoaD is not strongly sensitive to its hyperparameters, including the number of rollouts generated per scene, re-collecting CL data and further fine-tuning the policy, changing the number of optimization steps used for fine-tuning, or the number of trajectory samples ($K$) during CL data generation. In all alternative settings, RoaD improves upon the base model. 
While some hyper-parameter choices can further increase performance, in particular, increasing $K$ and re-collecting CL data for additional fine-tuning, these also increase the computational costs of RoaD.
On the other hand, we found that fine-tuning for too many optimization steps can slightly reduce performance, likely due to a lack of co-training with the original large-scale training data in our experiments. 

\myparagraph{Data scaling.} Given the high cost of scene reconstruction for E2E CL-SFT (i.e. generating 3D-GS artifacts), scalability of RoaD with the number of rollouts per scene is an important question. To this end, in \cref{fig:rollout_vs_score} we vary the number of rollouts generated per scene for CL-SFT. RoaD performance improves monotonically as more rollouts are added to its SFT dataset, while fine-tuning with re-simulated expert demonstrations cannot make use of multiple rollouts. 

\myparagraph{Sim2sim transfer.} Finally, in our experiments so far we have generated CL fine-tuning data in the same simulation environment where the policy is evaluated. Given a gap between simulation and the real-world, the policy may overfit to artifact of the simulation and in turn it may degrade in real-world deployment. While addressing sim2real gap is not in the scope of this work, to shed some light on this issue, we perform a sim2sim transfer experiment, where the policies are fine-tuned with 3DGS generated data, and evaluated in either 3DGS (default setting) or NeRF reconstructions (sim2sim transfer). For this experiment, we use an in-house scenarios set consisting of 75 scenarios cureted for dense ego-agent interactions. Results are reported in \cref{table:alpasim_sim2sim}. As expected, performance reduces when transferring fine-tuned policies to a new simulation environment, but fine-tuning with RoaD improves over the base model even in the transfer setting, indicating that RoaD has potential to improve real-world driving performance, despite possible sim2real gaps.

\section{Conclusions}
\label{sec:conclusions}
We presented RoaD, a simple closed-loop supervised fine-tuning (CL-SFT) method
that treats the policy’s own expert-guided rollouts as additional
demonstrations. By avoiding discrete recovery targets and introducing a
lightweight recovery mode, RoaD removes key assumptions that limit prior
CL-SFT approaches and makes the recipe applicable to modern E2E driving
policies, allowing closed-loop training without the need for reward functions.

Across vectorized traffic simulation and high-fidelity E2E driving,
RoaD consistently improves closed-loop performance over strong baselines. 
Because RoaD can achieve substantial improvements even when closed-loop data
is only collected once, it is a promising, data-efficient, approach for
training E2E driving policies in closed-loop.

Limitation of all CL-SFT approaches include the reliance of a pre-trained policy with sufficiently high performance, the assumption that the expert trajectory remains good behavior despite small deviations by the actor, and a distance metric for expert-guided rollouts. Further, our method relies on a high-fidelity simulator such as AlpaSim. Results on sim2sim transfer suggest that RoaD has potential to improve real-world driving performance, despite possible sim2real gaps. Future work may more explicitly address sim-to-real transfer and reduce overfitting to simulation, e.g., by co-training on simulated and real images, or introducing feature similarity bottlenecks~\cite{feng2025rap}.  

\clearpage
{
    \small
    \bibliographystyle{ieeenat_fullname}
    \bibliography{main}
}
\clearpage
\clearpage
\appendix
\maketitlesupplementary

\begin{figure*}[ht]
    \centering
    \includegraphics[width=\textwidth]{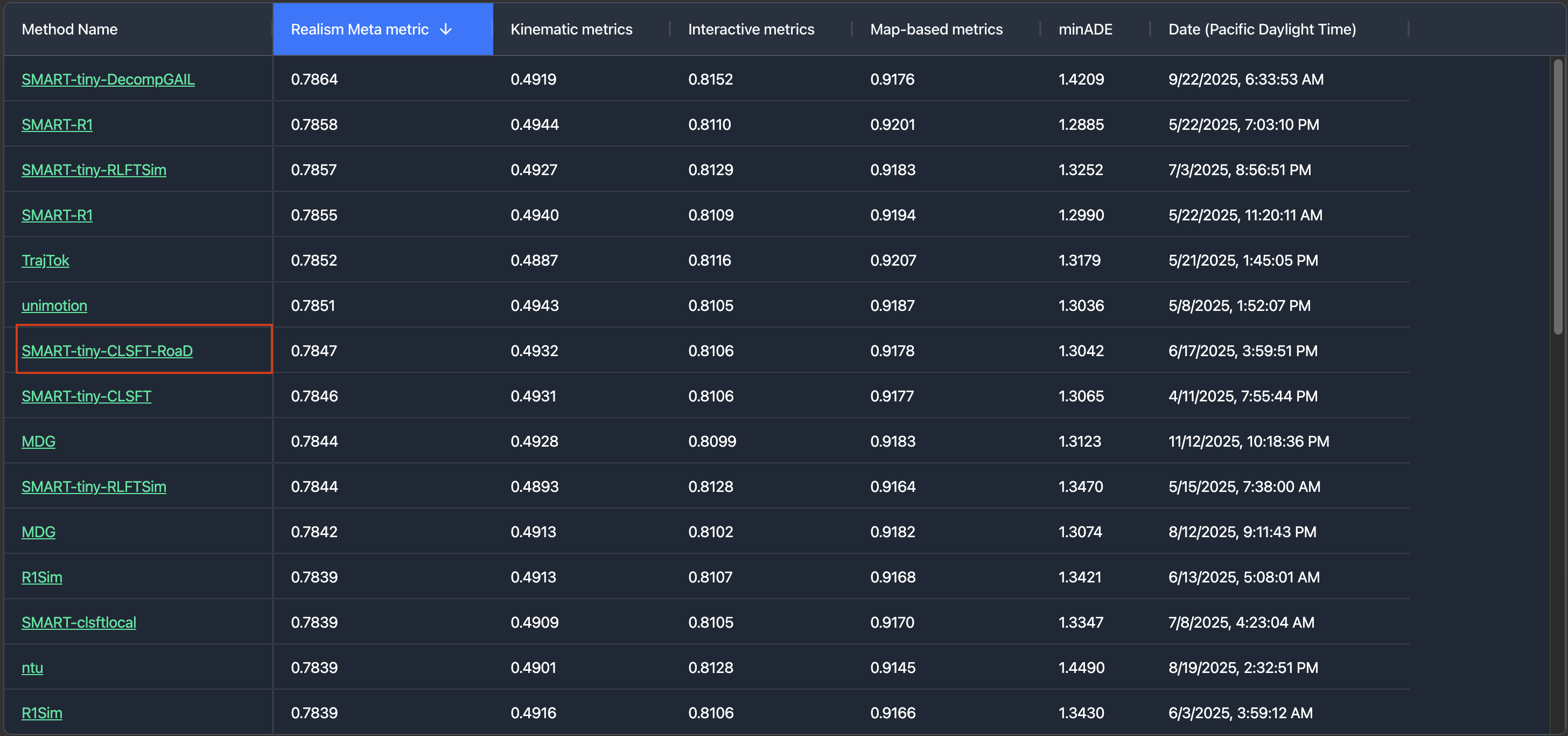}
    \caption{Snapshot of the WOSAC leaderboard with our SMART-tiny-CLSFT-RoaD entry highlighted (red box).}
    \label{fig:leaderboard-snapshot}
\end{figure*}

\section{Leaderboard Snapshot}

In \cref{fig:leaderboard-snapshot} we show a snapshot of the nuPlan WOSAC
leaderboard, with our \emph{SMART-tiny-CLSFT-RoaD} entry highlighted (red box).
We briefly discuss the other high-performing methods to clarify how they are
either orthogonal to our approach or specialized to the WOSAC task, and thus do
not directly transfer to E2E driving.

\emph{SMART-tiny-DecompGAIL} \citep{guo2025decomp}, \emph{SMART-tiny-RLFTSim}
\citep{ahmadi2024rlftsim}, and \emph{SMART-R1} \citep{pei2025advancing} use
reinforcement learning (RL) to optimize policies for the WOSAC task of matching
the data distribution. \emph{SMART-tiny-DecompGAIL} does so by using GAIL with
PPO, while \emph{SMART-tiny-RLFTSim} and \emph{SMART-R1} directly optimize the
WOSAC metric using RL (with small differences in implementation). However, these
methods do not translate well to E2E driving, where we typically lack a
well-defined reward function and RL tends to be too data-inefficient given the
high cost of high-fidelity simulation.

By contrast, \emph{TrajTok} \citep{zhang2025trajtok} proposes an improved
tokenizer for traffic models. This contribution is orthogonal to our approach.
However, it is only applicable to tokenizing short actions (e.g., one-step
actions), and hence does not translate to E2E driving, where policies typically
predict multiple seconds into the future.

Finally, \emph{unimotion} \citep{lin2025revisit} proposes an alternative to the
CatK rollout approach, whereby it finds the closest action to the ground truth
not only among the top-$K$ actions, but among all actions. As a result, it does
not require additional recovery actions (since it already tracks the
ground-truth trajectory as closely as possible), which yields a setup more
similar to our RoaD approach, where rollouts are taken directly as
demonstrations. However, this exhaustive search makes the method unsuitable for
E2E driving: it requires predicting and evaluating \emph{all} possible actions
of the policy, which is only feasible for small action spaces. This excludes
multi-token trajectory predictions, whose action space grows exponentially with
the horizon length, and flow-matching policies, whose action space is
continuous.
In their work, focussing on traffic models, they use either a fixed set of up to
2024 actions or a set of up to 16 actions predicted from action queries. 

\section{Experimental details: end-to-end driving}

\subsection{Simulation}

For data generation, we run AlpaSim~\cite{nvidia2025alpasim} at 30 Hz to match the frequency of the
ground-truth logs and reuse the same dataloader as for pre-training.
For evaluation, we run the simulator at 10 Hz, which matches the model’s training frequency.

At each step, we render two camera views (front-facing wide-angle and telephoto).
The policy predicts 6.4s trajectories, which are tracked by a downstream controller.
As in the main text, the controller models a 200 ms control delay and ego-motion noise,
and uses a dynamically extended bicycle model for the ego-vehicle dynamics.

Scenes are reconstructed using 3D Gaussian Splatting (3D-GS). Reconstruction quality
degrades with distance from the recorded trajectory. This is negligible for rollout
generation, where expert-guided rollouts remain close to the log, but can reduce
visual fidelity during evaluation if the ego vehicle deviates too far. To avoid such
artifacts, we discard rollouts whose ego trajectory deviates by more than 4 m from
the recorded trajectory.

All other traffic participants, including vehicles and pedestrians, replay their logged
trajectories and do not react to the ego vehicle. Consequently, they cannot avoid rear-end
collisions if the ego drives more slowly than in the recording. Following prior work
\citep{nuplan,cao2025pseudo,wang2025alpamayo}, we therefore count only ``at-fault'' collisions for the ego:
rear-end collisions caused by following vehicles are ignored, while lateral collisions are
still included.

\subsection{RoaD fine-tuning}

For RoaD rollout generation, we sample $K{=}64$ candidate trajectories from the
policy at a temperature of 0.8. To select the executed trajectory, we compute
the ground-truth distance $d^{\mathrm{g}}$ as the average distance between the
four corners of the ego vehicle along the predicted and ground-truth trajectories
over the first 20 time steps (2s). The same distance metric is used to decide
whether to trigger the recovery mode, with a threshold of $\delta_{\text{rec}}{=}3$ m.
When recovery is triggered, we linearly interpolate between the predicted and
ground-truth trajectories over $N_{\text{rec}}{=}30$ time steps and follow the
ground-truth trajectory thereafter. Recovery is disabled in the last 4s of the
episode because the controller requires at least 4s of input trajectory.

\end{document}